\def\year{2020}\relax
\documentclass[letterpaper]{article} 
\usepackage{aaai20}  
\usepackage{times}  
\usepackage{helvet} 
\usepackage{courier}  
\usepackage[hyphens]{url}  
\usepackage{graphicx} 
\usepackage{graphicx}  
\newcommand{\mname}{\texttt{ConCare}\xspace}

\usepackage{enumitem}
\usepackage{array}
\usepackage{multirow}
\usepackage{subfig}
\usepackage{scrextend}
\usepackage{xspace}
\usepackage{xcolor}
\usepackage{amsmath}

\title{$\mname$: Personalized Clinical Feature Embedding via Capturing the Healthcare Context}
\author{Liantao Ma\textsuperscript{\rm 1,\rm 3}, Chaohe Zhang\textsuperscript{\rm 1,\rm 3}, Yasha Wang \textsuperscript{\rm 1,\rm 2}\thanks{Corresponding Author} , Wenjie Ruan\textsuperscript{\rm 4}, Jiangtao Wang\textsuperscript{\rm 4}, \\ \Large \textbf{Wen Tang\textsuperscript{\rm 5}, Xinyu Ma\textsuperscript{\rm 1, \rm 3}, Xin Gao\textsuperscript{\rm 1,\rm 3}, Junyi Gao\textsuperscript{\rm 1, \rm 2}}
\\ 
\textsuperscript{\rm 1}Key Laboratory of High Confidence Software Technologies, Ministry of Education, Beijing, China\\ 
\textsuperscript{\rm 2}National Engineering Research Center of Software Engineering, Peking University, Beijing, China\\ 
\textsuperscript{\rm 3}School of Electronics Engineering and Computer Science, Peking University, Beijing, China\\ 
\textsuperscript{\rm 4}School of Computing and Communications, Lancaster University, UK\\ 
\textsuperscript{\rm 5}Division of Nephrology, Peking University Third Hospital, Beijing, China    
\\
\{malt, wangyasha\}@pku.edu.cn, \{wenjie.ruan, jiangtao.wang\}@lancaster.ac.uk, tanggwen@126.com
}
 
\begin{document}

\maketitle

\begin{abstract}
Predicting the patient's clinical outcome from the historical electronic medical records (EMR) is a fundamental research problem in medical informatics.
Most deep learning-based solutions for EMR analysis concentrate on learning the clinical visit embedding and exploring the relations between visits. Although those works have shown superior performances in healthcare prediction, they fail to explore the personal characteristics during the clinical visits thoroughly. Moreover, existing works usually assume that the more recent record weights more in the prediction, but this assumption is not suitable for all conditions. In this paper, we propose \mname to handle the irregular EMR data and extract feature interrelationship to perform individualized healthcare prediction. Our solution can embed the feature sequences separately by modeling the time-aware distribution. \mname further improves the multi-head self-attention via the cross-head decorrelation, so that the inter-dependencies among dynamic features and static baseline information can be effectively captured to form the personal health context. 
Experimental results on two real-world EMR datasets demonstrate the effectiveness of \mname. The medical findings extracted by \mname are also empirically confirmed by human experts and medical literature.
\end{abstract}

\section{Introduction}

\begin{figure}[]
\centering
\includegraphics[width=0.95\columnwidth]{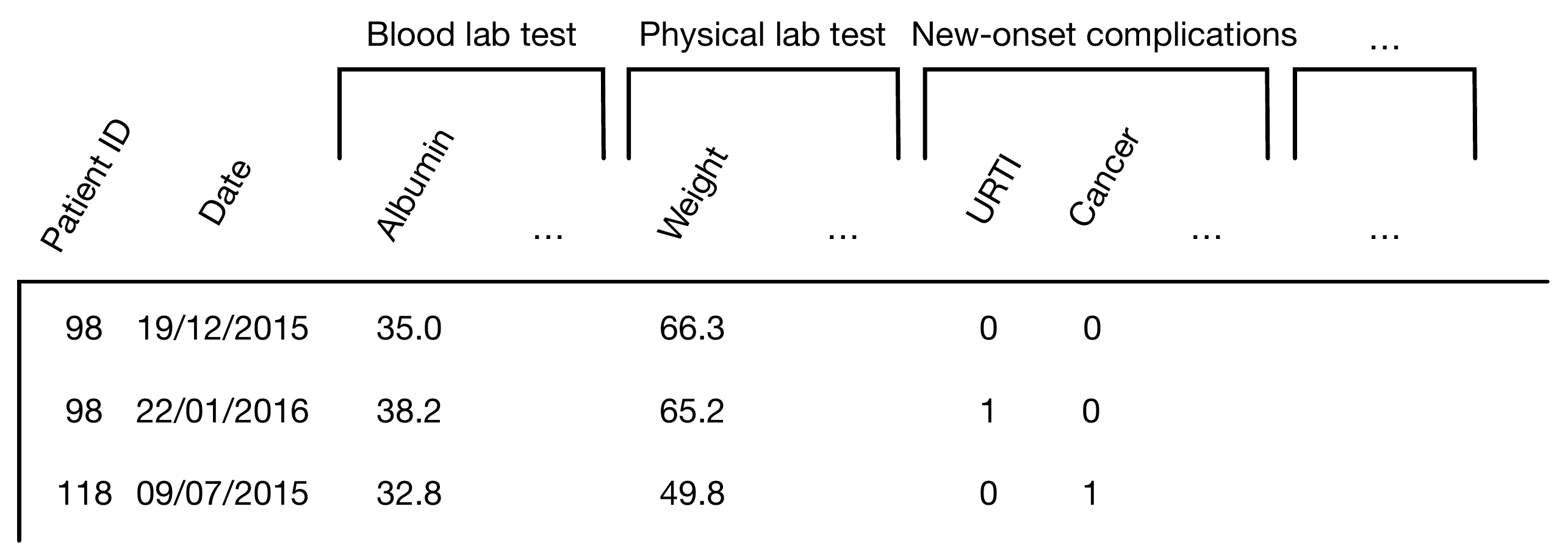}
\caption{Dynamic medical features. The physician conducts the necessary lab tests for the patient at each visit.}
\label{fig:dynamic_data}
\end{figure}

\begin{figure}[]
\centering
\includegraphics[width=0.95\columnwidth]{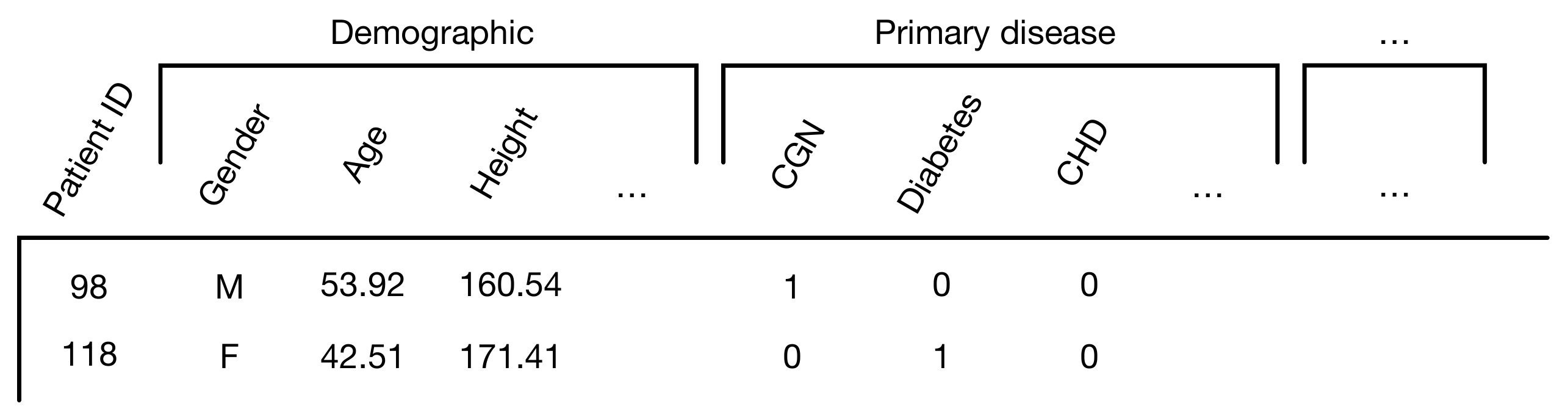}
\caption{Static baseline information. Such characteristics are usually used to evaluate the basic condition of the patient and the prognosis in the clinical practice.}
\label{fig:static_data}
\end{figure}

Performing personal health evaluation for each individual patient is always the goal that physicians pursue. 
Electronic Medical Records (EMR) now provide the possibility to realize these goals. 
EMR is a type of multivariate time series data that records patients' visits in hospitals (e.g., diagnoses, lab tests. As shown in Figure~\ref{fig:dynamic_data}) and static baseline information (e.g., gender, primary disease. As shown in Figure~\ref{fig:static_data}).
Recently 
deep learning-based models have demonstrated state-of-the-art performance in mining the massive EMR data~\cite{ma2020adacare,lee2017big,gaocamp2019,liu2018learning,liu2019learning}.
%
Usually, existing works incorporate multiple dynamic features (e.g., lab test values) to learn the visit embedding and the health status through the entire clinical visits by sequential models~\cite{ma2017dipole}.
%
%
Although the state-of-the-art performance has been demonstrated in these works,
the personal characteristics through clinical visits have not yet been fully taken into consideration on the healthcare prediction. Specifically, there are two research challenges, i.e., how to extract the different meanings of the particular clinical features for patients in diverse conditions, 
and how to evaluate the impact of irregular visit time intervals in the healthcare prediction.

\begin{itemize}[leftmargin=*]
\item\textbf{$I_1$: Extracting Personal Health Context}:
A certain value of a clinical feature (e.g., blood glucose) may imply different meanings to patients with diverse static baselines (e.g., diagnosis of diabetes as a primary disease, as shown in Figure~\ref{fig:static_data}).
In order to evaluate the health status of the patient comprehensively, physicians need to take a look at the static clinical baseline information.
Besides, not only the static baseline information, but also the dynamic clinical feature sequence (as shown in Figure~\ref{fig:dynamic_data}) can be treated as the health context of the patient.
For example, when plasma concentrations of creatinine and urea begin a hyperbolic rise, both of them and the GFR (i.e., Glomerular Filtration Rate) value are usually associated with systemic manifestations (uremia) for patients with chronic kidney disease \cite{msdmanuals_ckd}.
Thus, considering the particular condition of the patient, the way of attending the medical features in the whole prediction process should be individualized.
Some existing works try to model the relationship between clinical visits~\cite{ma2017dipole}, dynamic features~\cite{bai2018interpretable,choi2016retain} or incorporate the static information \cite{lee2018diagnosis}.
However, none of the existing models explored the interdependencies among dynamic records as well as static baseline information via a global view. 
In practice, it is critical to explore the inherent relationship between clinical features to build the personal healthcare context and perform the prediction individually.


\item\textbf{$I_2$: Capturing the Impact of Time Interval}:
The patient goes to the hospital only when he feels sick, and the physician prescribes lab examinations when it is necessary.
Therefore, medical records are produced irregularly in clinical practice.
It is assumed by many existing works~\cite{pham2016deepcare,baytas2017patient,ma2018health,bai2018interpretable} that the more recent clinical records weight more than previous records in general on the healthcare prediction.
However, under certain circumstance, historical records also contain valuable clinical information, which may not be revealed in the latest record (e.g., the blood glucose level was extremely abnormal). 
For instance, the upper respiratory tract infection (URTI) record a few years ago has almost no influence on the current healthcare prediction. However, the historical diagnosis of cerebrovascular disease indicates that the patient has been suffering from chronic cerebrovascular damage, so it is continuously the risk factor during the rest of the life~\cite{somers2008sleep}. 
Thus building a more adaptive time-aware mechanism to flexibly learn the impact of the time interval for each clinical feature is urgently needed.
\end{itemize}

To jointly tackle the above issues, in this paper, we propose a multi-channel healthcare predictive model, which can learn the representation of health status and perform the health prediction by more deeply considering the personal health context.
\mname evaluates the health status of patients mainly from the perspective of clinical features, rather than visits.
It embeds the time series of each feature separately.
The time decay effects of different features can be extracted separately and flexibly via corresponding learnable time-aware parameters.
The model explicitly extracts the interdependencies among time series of dynamic features as well as static baseline information, to learn the personal health context of patients in a global view.
\mname re-encodes each feature by looking at other features for clues that can help lead to a better understanding for this feature, so as to depict the health status more individually.
Our main contributions are summarized as follows:

\begin{itemize}[leftmargin=*]

\item
We propose a novel health status representation framework called \mname by fully considering the personal patient's health context. 
The health context is formed by capturing the interdependencies between clinical features which are extracted separately.  
To the best of our knowledge, we are the first research to jointly consider static baseline information, sequential dynamic features and the impact of the time interval as personal health context in the clinical representation learning.

\item 

Specifically, 
1) We explicitly extract interdependencies between clinical features to learn the personal health context and regenerate the feature embedding under the context, by a multi-head self-attention mechanism (addressing \textbf{$I_1$}). 
The cross-head decorrelation is utilized to
encourage the diversity among heads.
2) We propose a multi-channel medical feature embedding architecture, which learns the representation of different feature sequences via separate GRUs, and adaptively captures the effect of time intervals between records of each feature by time-aware attention (addressing \textbf{$I_2$}).

\item
We conduct the mortality prediction task on two real-world datasets (i.e., MIMIC-III dataset and end-stage renal disease dataset) respectively to verify the performance.
The results\footnote{We release our code and case studies at GitHub \url{https://github.com/Accountable-Machine-Intelligence/ConCare}} show that $\mname$ significantly and consistently outperforms the baseline approaches in both tasks.
We also reveal several interesting medical implications.
1) We provide the overall time-decay ratios for diverse biomarkers by the learnable parameter in time-aware attention.
2) We provide the adaptive cross-feature interdependencies, which further suggests possible medical research between specific features.
The obtained medical knowledge has been positively confirmed by human experts and clinical literature.


\end{itemize}

\section{Related Work}

\subsection{Exploring Relationship Among Clinical Records}

Most existing works only focus on exploring the relationship between clinical visits, in similar ways as general time series analysis and natural language processing tasks.
For example, Dipole \cite{ma2017dipole} uses bidirectional RNN architecture and the attention mechanism to capture the relationships of different visits for the prediction.
SAnD \cite{song2018attend} employs self-attention mechanism, positional encoding, and dense interpolation strategies to incorporate temporal order on clinical prediction tasks.

There are also a few novel research works that try to model the relationship between dynamic features rather than just the visits.
For example, \cite{bai2018interpretable} uses the self-attention mechanism to combine all diagnosis records produced in the visit to form the visit embedding, but it fails to extract the relationship in a global sequential view.  
\cite{gupta2018using} embeds the feature sequences by a pre-trained TimeNet, which cannot capture the unique characteristics for different features, respectively.
RETAIN \cite{choi2016retain} employs two RNNs to learn time attention as well as feature attention, and then sums up the weighted visit embedding to perform the prediction, but it lacks advanced feature extraction, and its prediction accuracy is limited \cite{ma2018health,ma2018risk}.

Besides the utilization of dynamic sequential data, several novel solutions also try to incorporate the static baseline information.
For example, \cite{lee2018diagnosis} proposes a medical context attention-based RNN that utilizes the derived individual information from conditional variational autoencoders. 
However, none of them explore the interdependencies between static baseline information and dynamic records from a global view.
The proposed model in this paper, \mname, can adaptively capture the relationship between clinical features and perform individualized prediction for patients in diverse health contexts.

\subsection{Handling the Time Interval between Visits}

Although most of the existing works \cite{ma2018kame,choi2017gram} simply treat the clinical visits in equal intervals, 
several novel works \cite{pham2016deepcare,ma2018health} try to model the importance of clinical visits with time intervals, by attaching a fixed time-decay ratio to decay the hidden memory of the previous visit. Those works omit the different characteristics between features.
For example, 
T-LSTM \cite{baytas2017patient} handles irregular time intervals of visits in longitudinal patient records by enabling time decay to discount the cell memory content in LSTM. 
In order to capture the characteristics of different disease codes, Timeline \cite{bai2018interpretable} develops time decay factors for diseases to form visit representation and feeds it into an RNN for prediction. 
But its time-aware effect is still disrupted during the historical visit embedding process due to the rapid memory forgetting of RNN.
And Timeline can only handle the disease codes as features rather than biomarkers.

Therefore, existing works simply assume that recent records play more important roles than previous records.
However, according to the clinical practice, some historical clinical events also strongly indicate the health status under certain circumstances while it may not be revealed in the latest record.
The time-aware mechanism should take the characteristics of features into consideration and meanwhile, flexibly retain the vital historical information.
\mname can capture the impact of time interval in diverse feature sequences by a learnable time-aware attention.

\section{Problem Formulation}

We assume that the patient's dynamic clinical records (as shown in Figure~\ref{fig:dynamic_data}) consist of $T$ visits to the hospital.
The number of features in each visit record is $N$.
As a result, such a clinical sequence can be formulated as a “longitudinal patient matrix” $record$ where one dimension represents medical features and the other one denotes visit timestamps~\cite{lee2017big}:

\begin{equation}
\textbf{$record$}=\begin{pmatrix}
r_{11} & \cdots & r_{1T} \\
\vdots & \ddots & \vdots \\
r_{N1} & \cdots & r_{NT}\\
\end{pmatrix}.
\end{equation}

The static baseline data (as shown in Figure~\ref{fig:static_data}), including demographic attributes and historical primary diseases,
is denoted as $base$. 
The objective of healthcare prediction is using EMR data (i.e., $record$ and $base$) to predict whether a patient suffers from the target health risk during the period of the treatment procedure, denoted as $y$ $\in \left \{ 0, 1 \right \}$. %
This problem is posed as a binary classification
under a certain time window (e.g., 24 hours), namely, $\widehat{y} = \mname(record, base).$

\section{Solution}
\begin{figure}[t]
\centering
\includegraphics[width=0.95\columnwidth]{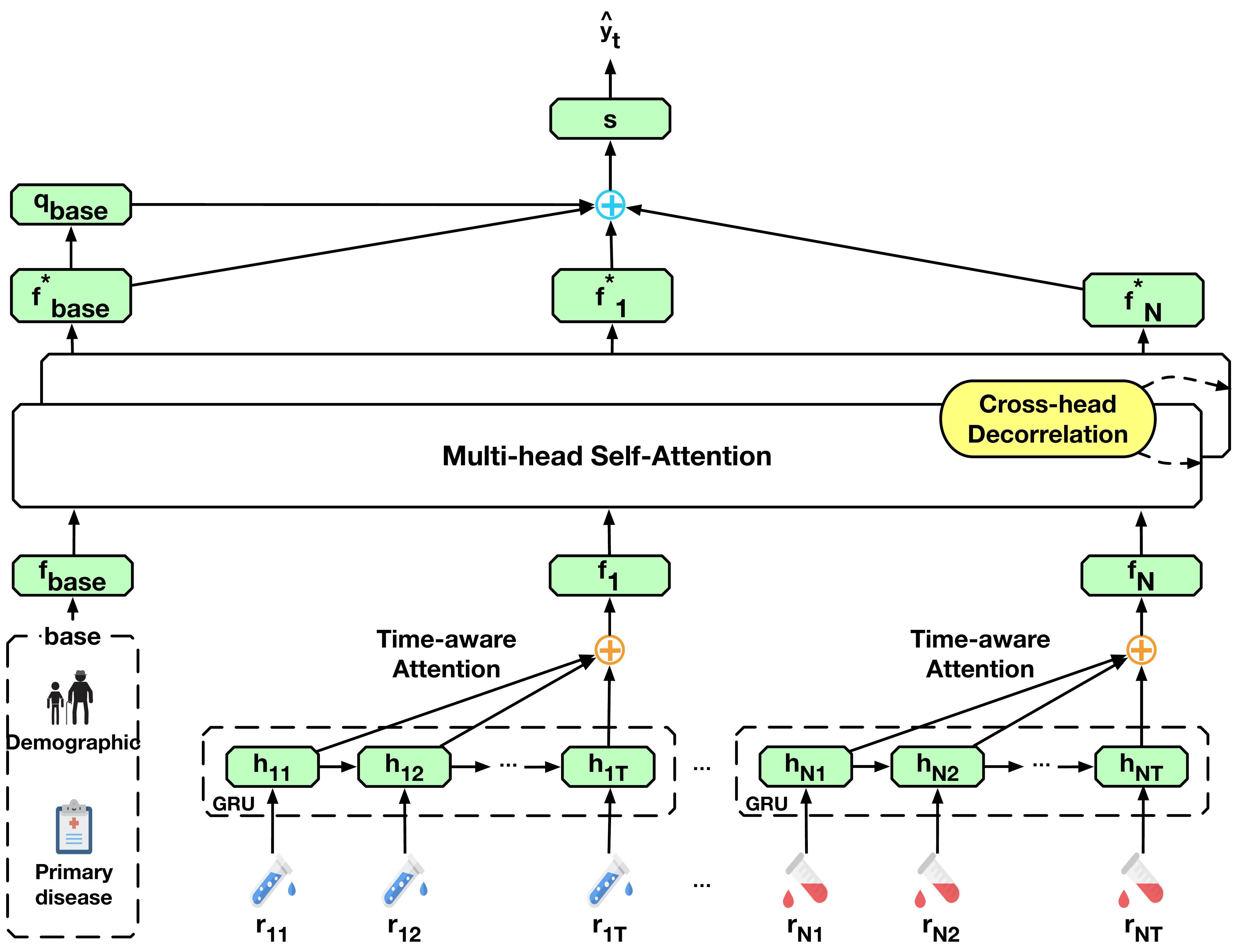}
\caption{The Framework of \mname.}
\label{fig:framework}
\end{figure}

Figure \ref{fig:framework} shows the framework of the proposed \mname. 
The model treats the clinical information of the patient from the perspective of features rather than visits.
We extract the context vector of each dynamic feature and static baseline information separately.
Such feature embedding vector are then re-encoded by taking the information of all features as healthcare context.
The framework comprises of the following sub-modules:

\begin{itemize}
    \item The multi-channel time series embedding module with time-aware attention is developed to separately learn the representation of each dynamic feature.
    \item The feature encoder is adopted to combine all the static information and dynamic records based on self-attention.
\end{itemize}

The individualized prediction finally is obtained from all regenerated feature embeddings with an attention queried by static baseline information.
We will present the details in the following subsections.

\subsection{Multi-Channel Clinical Sequence Embedding}

In \mname, we aim to capture the interdependencies between features based on self-attention mechanism~\cite{vaswani2017attention}.
Since the self-attention architecture contains no recurrence, in order to incorporate information about the order of the sequence, researchers simply utilize the fixed positional encoding to provide relative position information for timestamps~\cite{song2018attend}.
However, such positional embedding capability is limited, especially for the absolute position understanding, but the logical order of the clinical sequence actually matters in the medical domain.
\mname thus embeds the time series of each feature separately by multi-channel GRU:

\begin{equation}
h_{n,1}, ..., h_{n,T} = GRU_n(r_{n, 1}, ..., r_{n, T}),
\end{equation}
where, the time series of feature $n$ is denoted as 
 $r_{n,:} = (r_{n,1}$, ...,$ r_{n,T} )$ 
$\in {R}^{T}$.
Then, 
the hidden representations is summarized across the whole time span. 
To capture the impact of time intervals in each sequence, we propose a time-aware attention mechanism here.
Generally, an attention function can be described as mapping a query and a set of key-value pairs to an output \cite{vaswani2017attention}. 
The output is computed as a weighted sum of the values, where the weight assigned to each value is computed by a compatibility function of the query with the corresponding key. 
First, the $Query$ vector is generated by the hidden representation at the last time step $T$, and the $Key$ vectors are generated by each hidden representation:
\begin{equation}
q^{emb}_{n,T} = W^{q}_{n} \cdot h_{n,T},
\end{equation}
\begin{equation}
k^{emb}_{n,t} = W^{k}_{n} \cdot h_{n,t},
\end{equation}
where $q^{emb}_{n,T} $ and $k^{emb}_{n,t} $ are the $Query$ vector and the $Key$ vector respectively. 
$W^{q}_{n} $ and $W^{k}_{n} $ are the corresponding projection matrices to obtain the query and key vectors. 
Then we design the time-aware attention weights as follow:
\begin{equation}
\alpha_{n,1}, \alpha_{n,2}, ..., \alpha_{n,T} = Softmax(\zeta_{n,1}, \zeta_{n,2}, ..., \zeta_{n,T}),
\end{equation}
where
\begin{equation}
\zeta_{n,t} = tanh( \frac{q^{emb}_{n,T} \cdot k^{emb}_{n,t}} {\beta_{n} \cdot log(e+(1-\sigma(q^{emb}_{n,T} \cdot k^{emb}_{n,t}))\cdot\Delta t)}  ).
\label{eq:cal_atten}
\end{equation}
This is an alignment model that can quantify how much each hidden representation contributes to the densely summarized representation for each feature. 
$\Delta t$ is the time interval to the latest record. 
$\sigma$ is the $sigmoid$ function.
$\beta_{n}$ is a feature-specific learnable parameter trained to control the influence of the time interval on the corresponding feature. 
The attention weight $\alpha_{n,t}$ will be significantly decayed, if:

\begin{itemize}
    \item the time interval $\Delta t$ is long, which means that such value is recorded a long time ago. It is obviously that, the most recent (i.e., $\Delta t = 0$) value of any feature will only be decayed slightly (i.e., $log(e) = 1$).
    
    \item the time-decay ratio $\beta_{n}$ is high which means that for particular clinical feature only recent recorded value matters. The clinical feature whose influence persists (i.e., $\beta_{n}$ is low) will be decayed just slightly.

    \item the historical record does not actively respond to the current health condition (i.e., $q^{emb}_{n,T} \cdot k^{emb}_{n,t}$ is small).
\end{itemize}

Finally, based on the learned weights, we can derive time-aware contextual feature representation as 
$f_{n} = \sum_{i=1}^{T} \alpha_{n,t}\cdot h_{n,t}.$
Furthermore, the demographic baseline data is embedded into the same hidden space of $f_{n}$
\begin{equation}
f_{base} = W^{emb}_{base} \cdot base,
\end{equation}
where $W^{emb}_{base}$ is an embedding matrix. 
Thus, all the data of the patient can be represented by a matrix $F$ (i.e., a sequence of vectors, where each vector represents one feature of the patient over time):
$F=(f_{1},\cdots,f_{N},f_{base})^{\top}$.

\subsection{Learning the Context and Re-encoding the Feature}


We capture the interdependencies among dynamic features through visits as well as static baseline information, and further re-encode the feature embedding under the personal context based on self-attention.
As the \mname processes each feature, self-attention allows it to look at other features for clues that can help lead to a better encoding for this feature.
For example, when the model is processing the feature “blood glucose”, self-attention may allow it to associate it with “diagnosis of diabetes” in the static baseline information.
Besides, the multi-head mechanism enhances the attention layer with multiple representation subspaces.
Mathematically, given current feature representations $F$, the refined new representations are calculated as:
\begin{equation}
\begin{split}\label{eq:head_concat}
u_{n} &= MultiHeadAttention(F)\\
& = [head_{1}(f_{n})\oplus head_{2}(f_{n}) \oplus ...\oplus head_{m}(f_{n})]W^{O},
\end{split}
\end{equation}
where $head_{m}$ is $m$-$th$ attention head, $\oplus$ is the concatenation operation and $W^{O}$ is a linear projection matrix.
Considering both efficiency and effectiveness, the scaled dot product is used as the attention function \cite{vaswani2017attention}.
This following $softmax$ score determines how much each feature will be expressed at this certain feature.
Specifically, $head_{m}$ is the weighted sum of all value vectors and the weights are calculated by applying attention function to all the query, key pairs:

\begin{equation}
\begin{split}\label{eq:scale_dot}
\alpha_{1}, \alpha_{2}, ..., \alpha_{N+1} = \text{Softmax}(\frac{q\cdot k_{1}}{\sqrt{d_k}}, \frac{q\cdot k_{2}}{\sqrt{d_k}}, ..., \frac{q\cdot k_{N+1}}{\sqrt{d_k}}),
\end{split}
\end{equation}
\begin{equation}
\begin{split}\label{eq:sum_up}
head_{m}(g_{n}) = \sum_{i=1}^{N+1} \alpha_{i}\cdot v_{i},
\end{split}
\end{equation}
where $q$, $k_{i}$ and $v_{i}$ are the query, key, and value vectors and $d_k$ is the dimension of $k_{i}$. Moreover, $q$, $k_{i}$ and $v_{i}$ are obtained by projecting the input vectors into query, key and value spaces, respectively \cite{wang2019r}. 
They are formally defined as:
\begin{equation} 
\begin{split}
q, k_i, v_i = W^{q} \cdot f, W^{k} \cdot f_i, W^{v} \cdot f_i ,
\end{split}
\end{equation}
where $W^{q}$, $W^{k}$ and $W^{v}$ are the projection matrices and each $head_{m}$ has its own projection matrices. 
As shown in Eq.\ref{eq:scale_dot} and Eq.\ref{eq:sum_up} each $head_{m}$ is obtained by letting $f$ attending to all the $fearure$ positions, thus any feature interdependencies between $f$ and $f_i$ can be captured.

\subsection{Cross-Head Decorrelation}

Heads for self-attention are expected to capture dependencies from different aspects.
However, in practice, heads may tend to learn similar dependencies according to \cite{vaswani2017attention}.
To overcome this challenge, 
we encourage diverse or non-redundant representations \cite{cogswell2015reducing,ChuMLRDA2019} by minimizing the cross-covariance of hidden activations across different heads. 
We utilize the cross-head decorrelation module to expand the model’s ability to focus on different features, based on \cite{cogswell2015reducing} which reduces the redundancy of the normal neural network layer.
According to Eq.\ref{eq:head_concat}, we get $u_t$ as the multi-head attention for $f_t$, which is the concatenation of the heads. 
For simplicity, here we use $u$ to denote $u_t$ as a general case.
The covariances between all pairs of activations $i$ and $j$ of $u$ form a matrix $C$:
\begin{equation}
C_{i,j} = \frac{1}{B}\sum_{b = 1}^{B}(u^b_i-\mu_i )(u^b_j-\mu_j ),
\end{equation}
where $B$ is the batch size and $u^b_i$ is the $i-th$ activation of $u$ at $b-th$ case in the batch.
$\mu_i = \frac{1}{B} \sum_{b = 1}^{B} u^b_i $ is the sample mean of activation $i$ over the batch.
Covariance between diverse heads is expected to be minimized.
The diagonal of $C$ is then subtract from the matrix norm to build the cross-head decorrelation loss term:
\begin{equation}
\mathcal{L}_{decorrelation} = \frac{1}{2}(\left \| C \right \|_F^{2} - \left \| diag(C) \right \|_2^{2}),
\end{equation}
where $\left \| \cdot \right \|_F$ is the frobenius norm,and the $diag()$ operator extracts the main diagonal of a matrix into a vector. 
After obtaining the refined representation of each position by the multi-head attention mechanism, we add a position-wise fully connected feed-forward network sub-layer. This feedforward network transforms the features non-linearly and is defined as 
$FeedForward(r_{n}) = max(0, u_{n} \cdot W_1 +b_1) \cdot W_2 + b_2$.
We also employ a residual connection \cite{he2016deep} around each of the two sub-layers, followed by layer normalization \cite{ba2016layer}.
As shown in Fig.\ref{fig:framework}, the outputs of this subsection from $F$ are denoted as $F^{*} = (f^{*}_{1}, f^{*}_{2}, ..., f^{*}_{N}, f^{*}_{base})^{\top}$.

\subsection{Healthcare Prediction}

A dense health status representation is expected to perform the final prediction.
Here, we introduce an individualized characterization attention summarization.
The $Query$ is obtained by $f^{*}_{base}$ and $Key$s are formed by $F^*$ as:
\begin{equation}
    q^{fin}_{base} = W^{fin}_{base} \cdot f^{*}_{base},
\end{equation}
\begin{equation}
    k^{fin}_{n} = W^{fin}_{n} \cdot f^{*}_{n},
\end{equation}
where $W^{fin}_{base}$ and $W^{fin}_{n}$ are the projection matrix respectively. 
Similar to the first subsection, the attention weights are calculated as:
\begin{equation}
\alpha^{fin}_{base}, \alpha^{fin}_{1}, ..., \alpha^{fin}_{N} = \text{Softmax}(\zeta^{fin}_{base}, \zeta^{fin}_{1}, ..., \zeta^{fin}_{N}),
\end{equation}
\begin{equation}
\zeta^{fin}_{n} = tanh( q^{fin}_{base} \cdot k^{fin}_{n}  ).
\end{equation}
The health status representation $s$ and the prediction result $\widehat{y}$ can be obtained by:
\begin{equation}
s = \sum_{i=1}^{N} \alpha^{fin}_{i}\cdot f^{*}_{i} + \alpha^{fin}_{base} \cdot f^{*}_{base},
\end{equation}
\begin{equation}
\widehat{y} = \sigma(W^{fin} \cdot s + b^{fin}),
\end{equation}
where $W^{fin}$ and $b^{fin}$ are the weight matrix and bias term, respectively. And the final loss can be denoted as the combination of cross-entropy loss and decorrelation loss
\begin{equation}
\mathcal{L} = \mathcal{L}_{cross-entropy} + \mathcal{L}_{decorrelation}.
\end{equation}

\section{Experiment}

We conduct the mortality prediction experiments on MIMIC-III dataset \footnote{https://mimic.physionet.org} and end-stage renal disease (ESRD) dataset.
The source code of \mname, statistics of datasets and case studies are available at the GitHub repository\footnote{\url{https://github.com/Accountable-Machine-Intelligence/ConCare}}.

\subsection{Datasets and Prediction Tasks}

\begin{itemize}
\item
\noindent \textbf{MIMIC-III Dataset.} We use ICU data from the publicly available Medical Information Mart for Intensive Care (MIMIC-III) database \cite{johnson2016mimic}.
We perform the in-hospital mortality prediction for patients 
based on patients' demographic data and events produced during ICU stays \cite{harutyunyan2017multitask}. 
We fix a test set of 15\% of patients and divide the rest of the dataset into the training set and validation set with a proportion of 0.85\,:\,0.15.

\item
\noindent \textbf{Real-World ESRD Dataset.} We perform the mortality prediction on an end-stage renal disease dataset.
The cleaned dataset consists of 656 patients with static baseline information and 13,091 dynamic records. 
There are 1196 records with positive labels (i.e., died within 12 months) and 10,804 records with negative labels.
The training set is further split into 10 folds to perform the 10-fold cross-validation.
\end{itemize}

We assess performance using the area under the receiver operating characteristic curve (AUROC), area under the precision-recall curve (AUPRC), and the minimum of precision and sensitivity Min(Se,P+).
AUPRC is the most informative and the primary evaluation metric when dealing with a highly imbalanced and skewed dataset \cite{davis2006relationship,choi2018mime} like the real-world EMR data.

\subsection{Implementation Details and Baseline Approaches}

The training was done in a machine equipped with CPU: Intel Xeon E5-2630, 256GB RAM, and GPU: Nvidia Titan V by using Pytorch 1.1.0. 
For training the model, we used Adam \cite{kingma2014adam} with the mini-batch of 256 patients and the learning rate is set to $1e-3$. 
To fairly compare 
different approaches, the hyper-parameters of the baseline models are fine-tuned
by grid-searching strategy.
%
We include several state-of-the-art models as our baseline approaches.

\begin{itemize}



\item GRU$_{\alpha}$ is the basic GRU with an addition-based attention mechanism.

\item RETAIN (NeurIPS 2016) \cite{choi2016retain} utilizes a two-level neural attention mechanism to detect influential visits and significant variables.

\item T-LSTM (SIGKDD 2017) \cite{baytas2017patient} handles irregular time intervals by enabling time decay. We modify it into a supervised learning model.

\item MCA-RNN (ICDM 2018) \cite{lee2018diagnosis} utilizes the derived individual patient information from conditional variational auto-encoders to construct a medical context attention-based RNN.

\item Transformer$_e$ (NeurIPS 2017) \cite{vaswani2017attention} is the encoder of the Transformer, in the final step, we use to flatten and FFNs to make the prediction.

\item SAnD$_{*}$ (AAAI 2018) \cite{song2018attend} models clinical time-series data solely based on masked self-attention. 
When performing prediction at every time step, we use causal padding \cite{van2016wavenet} for the convolutional layer to prevent using future information.
We re-implement SAnD by using $r_{t-k+1:t}$ to build input embedding at the measurement position $t$, instead of the one proposed in the original paper $r_{t:t+k-1}$, to avoid the violation of causality.

\end{itemize}

For a fair comparison, 
although most of the comparative approaches did not take the static baseline information into consideration which is greatly beneficial for improving the performance of healthcare prediction,
we feed such characteristics as additional input (i.e., concatenate with the raw input) for them at each visit.

\subsection{Results of Risk Prediction}

\begin{table*}[t]
  \centering
  \caption{Results of the Healthcare Prediction Tasks}
  \label{tab:result}

\begin{tabular}{ccccccc}
\hline

 & \multicolumn{3}{c}{MIMIC-III Dataset (Bootstrapping)} & \multicolumn{3}{c}{ESRD Dataset (10-Fold Cross Validation)} \\
 
 Methods & AUROC & AUPRC & min(Se, P+)  & AUROC & AUPRC & min(Se, P+) \\ 
\hline
 
 
 
 GRU$_{\alpha}$ & .8628\,(.011)  & .4989\,(.022) & .5026\,(.028) & .8066\,(.004)  & .3502\,(.009) & .3770\,(.006)   \\
 
 RETAIN & .8313\,(.014) & .4790\,(.020) & .4721\,(.022) &.7986\,(.005)  & .3386\,(.009) & .3699\,(.011)  \\

MCA-RNN  & .8587\,(.013) & .5003\,(.028) & .4932\,(.024) & .8021\,(.015)  & .3451\,(.041) & .3731\,(.025)  \\

T-LSTM  & .8617\,(.014)  & .4964\,(.022) & .4977\,(.029)  & .8101\,(.015)  & .3508\,(.052) & .3721\,(.045)  \\

Transformer$_{e}$ & .8535\,(.014)  & .4917\,(.022) & .5000\,(.019)  & .8082\,(.027)  & .3502\,(.062) & .3719\,(.037)  \\

SAnD$_{*}$  & .8382\,(.007) & .4545\,(.018) & .4885\,(.017) & .8002\,(.026)  & .3371\,(.036) & .3591\,(.053)  \\

 \hline
$\mname^{}_{PE}$ & .8566\,(.008) & .4811\,(.024) & .5012\,(.020) & .8124\,(.025)  & .3561\,(.047) & .3761\,(.037)  \\
$\mname^{}_{MC-}$ & .8594\,(.008) & .4902\,(.024) & .4947\,(.025) &.8101\,(.023)  & .3498\,(.066) & .3766\,(.064)  \\
$\mname^{}_{DE-}$ & .8671\,(.009) & .5231\,(.028) & .5080\,(.023) &.8162\,(.033)  & .3525\,(.063) & \textbf{.3864}\,(.034)  \\

 \hline
$\mname$ & \textbf{.8702}\,(.008) &\textbf{.5317}\,(.027) & \textbf{.5082}\,(.021) & \textbf{.8209}\,(.036) & \textbf{.3606}\,(.084) & .3853\,(.071)  \\

\hline
\end{tabular}
\end{table*}

Table \ref{tab:result} shows the performance of all approaches on two datasets.
The number in $()$ denotes the standard deviation of bootstrapping for 100 times on the MIMIC-III dataset and the standard deviation of 10-fold cross-validation on the ESRD dataset.
The results indicate that \mname significantly and consistently outperform other baseline methods.

We find that \mname outperforms the approaches that only utilize the embedding of the health status in visits (i.e., $\mname^{}_{MC-}$ and all comparative approaches). 
\mname also outperforms the approaches which incorporate the static information.
It indicates that capturing the interdependencies among clinical features (including static baseline information and dynamic features) and regenerating the feature embedding under the personal health context is critical for evaluating the health status. 
%

Moreover, \mname outperforms the positional encoding-based approaches (i.e., $\mname^{}_{PE}$, Transformer-Encoder, SAnD).
It demonstrates the superior of multi-channel GRU encoder than the conventional positional encoding which is difficult to precisely embed the positional information.
\mname also outperforms the time-aware approaches (i.e., T-LSTM), which demonstrates that capturing the time-decay impact of each feature separately in a global view is superior to directly decaying the hidden memory of entire visits.
The superior performance of \mname than the $\mname_{DE-}$ (i.e., without the decorrelation loss) verifies the efficacy of the decorrelation loss which can encourage the diversity among heads and improve the performance.

\subsection{Findings and Implications}
This section will discuss the findings and implications of $\mname$ in the experiments.
\subsubsection{Decay Rates For Different Features}

\begin{figure}[htb]
\centering
\includegraphics[width=0.95\columnwidth]{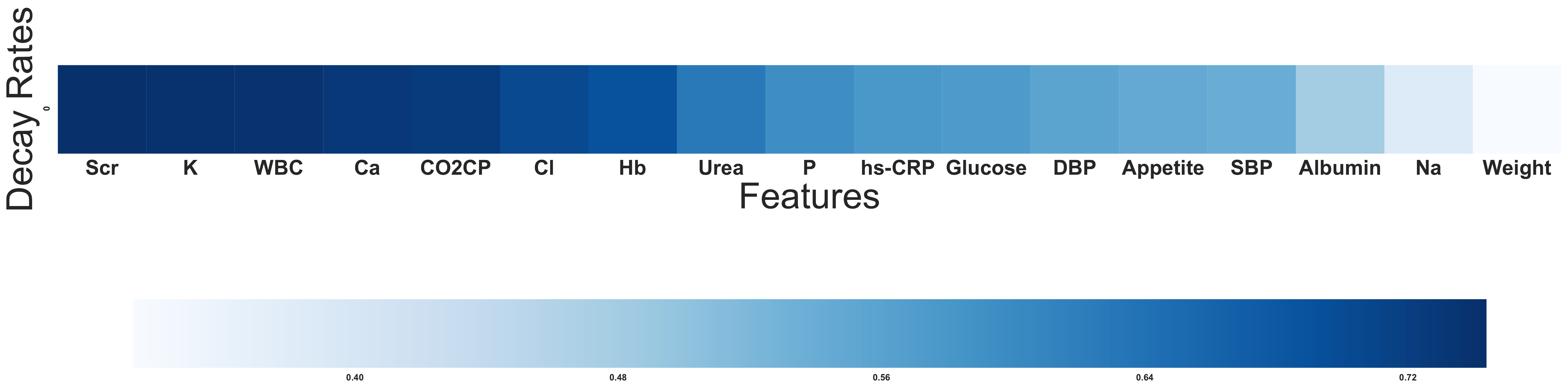}
\caption{Decay Rates For Different Features}
\label{fig:decay_rates}
\end{figure}
Figure \ref{fig:decay_rates} shows the decay rates (i.e. the $\beta$ in Eqn. \ref{eq:cal_atten}) learned adaptively for different features, which depicts how the importance of previous values of features fades through time. 
The darker boxes mean the importance of previous values of features fades quickly (i.e., the short-term patterns of the features matter), and vice versa.
The figure indicates that \mname attends more on the short-term of serum creatinine (Scr), K, White Blood Cell Count (WBC), Ca, Carbon-dioxide Combining Power (CO2CP), Cl, hemoglobin (Hb).
According to the medical commonsense, the above features are relatively fast-changing indicators, reflecting the patient's infection status or dialysis adequacy, etc. 
Conversely, the weight, albumin, Na and systemic blood pressure (SBP) need to be attended in the long-term aspect.
According to medical research~\cite{meijers2008review}, these features are usually related to nutrition intake and reflect the patient's condition over a period of time.

\subsubsection{Cross-Feature Interdependencies}

\begin{figure}[htb]
\centering
\includegraphics[width=1\columnwidth]{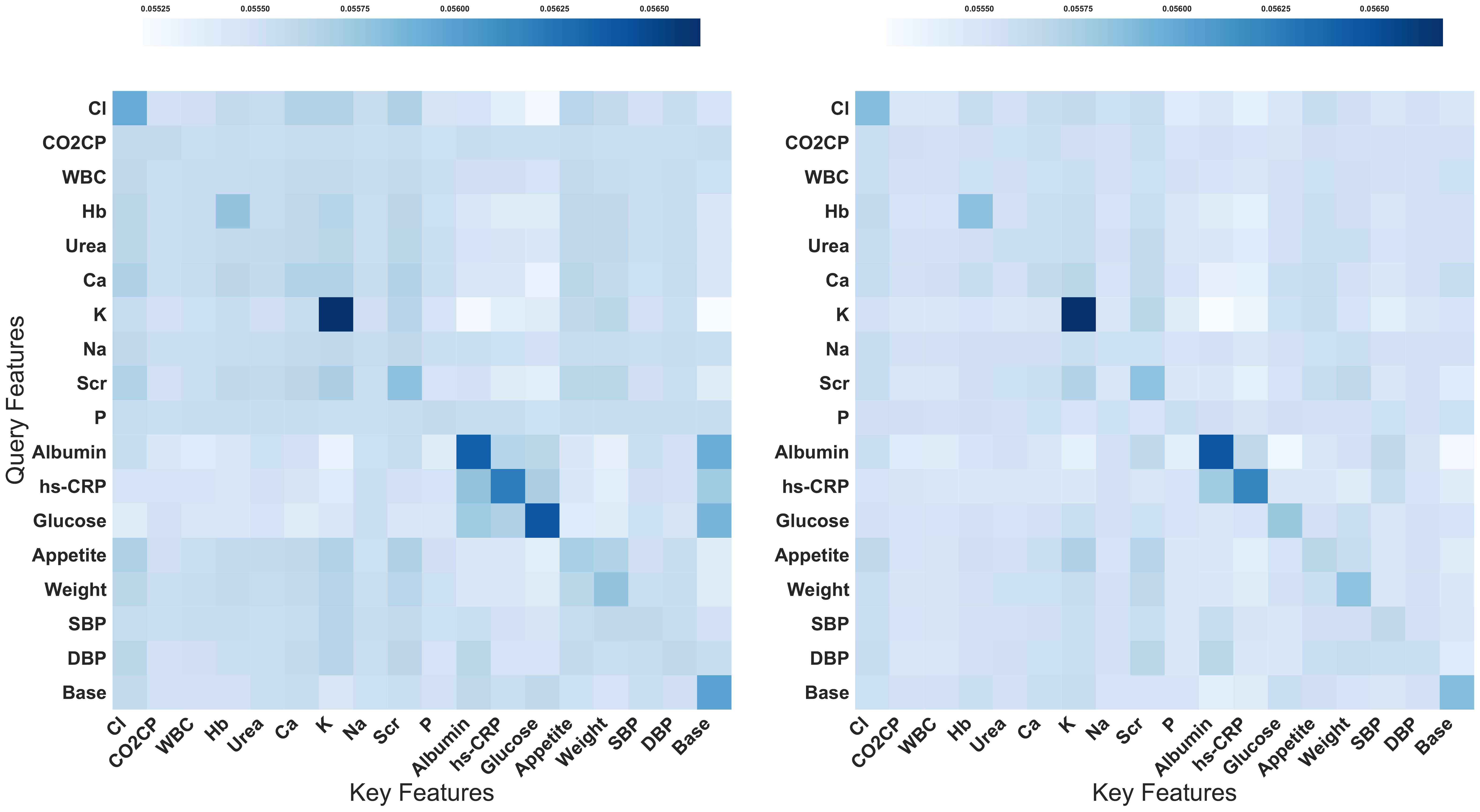}
\caption{Cross-Feature Interdependencies: Patients Died with (Left) / without (Right) Diabetes}
\label{fig:dia_studies}
\end{figure}

Figure~\ref{fig:dia_studies} shows cross-feature interdependencies of all patients who died with/without diabetes respectively.
The average attention weights of one head calculated by the self-attention module are shown.
The ordinates of the two figures are the $Query$ features and the abscissas are the $Key$ features.
The boxes in the figures show when a $Query$ feature makes a query, how much each $Key$ feature respond to the $Query$.
Most of the clinical features are more likely to respond to themselves, which denoted by the diagonal of two matrices.
It is common medical knowledge that the glucose of a patient is strongly related to diabetes.
By comparing the two figures, in the box of Glucose-Glucose position, the model pays much more attention to the glucose in patients who died with diabetes.
Besides, \mname figures out that there are relatively high interdependencies between albumin, hyper-sensitive C-reactive protein (hs-CRP), glucose and the static information (including age, diagnosis of diabetes) for patients suffering from diabetes.
This is highly consistent with the medical research~\cite{milan2015pro} and medical experience.

\section{Conclusion} 

In this work, we proposed a novel medical representation learning framework, \mname, which can explicitly extract the personal healthcare context and perform health prediction individually.
Specifically, it extracts the clinical features by multi-channel GRU with a time-aware attention mechanism.
The interdependencies among static baseline information and dynamic features are captured to build the health context and re-encode the clinical information.
We conducted experiments on two real-world datasets.
\mname demonstrated significant prediction performance improvement across both tasks.
It provides the time-decay ratios for different features respectively and indicates the interdependencies between features as interpretability.
All extracted medical findings have been positively confirmed by experts and medical literature. The results also remind some possible medical research opportunities for deeply analyzing the relationship between some clinical features.

\section{ Acknowledgments}
This work is supported by the National Science and Technology Major Project (No. 2018ZX10201002), and the fund of the Peking University Health Science Center (BMU20160584). WR is supported by ORCA PRF Project (EP/R026173/1).

\bibliographystyle{aaai}
\bibliography{5769_ref}

\end{document}